\documentclass[10pt,aps,prd,email,showpacs,preprintnumbers,showkeys,superscriptaddress,amsmath,amssymb]{revtex4-2}
\usepackage{pgfplots}
\usepackage{tikz}
\usetikzlibrary{positioning}
\usepackage{color} 
\usepackage{hyperref}
\usepackage{slashed}
\usepackage{graphicx}
\usepackage{latexsym}
\usepackage{epstopdf}
\usepackage{amsmath}
\usepackage{enumitem}
\usepackage{amssymb,amsmath}
\usepackage{multirow}
\usepackage{mathtools}
\usepackage{bbm}
\usepackage{bm}
\usepackage{amsfonts}
\usepackage{amssymb}
\usepackage{tabularx}
\usepackage{calligra}
\usepackage{calrsfs}
\usepackage{makecell}
\usepackage[normalem]{ulem}
\usepackage[USenglish,american]{babel}
\usepackage{xcolor}
\usepackage{hyperref}
\hypersetup{
  colorlinks=false,             
  linkbordercolor=blue,         
  citebordercolor=green,
  urlbordercolor=blue,
  pdfborder={0 0 1}             
}
\usepackage{subcaption}
\usepackage[compat=1.1.0]{tikz-feynman}
\renewcommand{\arraystretch}{1.8}

\expandafter\ifx\csname package@font\endcsname\relax\else
 \expandafter\expandafter
 \expandafter\usepackage
 \expandafter\expandafter
 \expandafter{\csname package@font\endcsname}%
\fi
\begin{document}
\title{Neuro-evolutionary stochastic architectures in gauge-covariant neural fields}

\author{Rodrigo Carmo Terin} 
\email{rodrigo.carmo@urjc.es}
\affiliation{King Juan Carlos University, Facuty of Experimental Sciences and Technology, Department of Applied Physics, Av. del Alcalde de Móstoles, 28933, Madrid, Spain}

\begin{abstract}
We extend our gauge-covariant stochastic neural-field framework by promoting architecture-level parameters to slow stochastic variables evolving in function space. Our effective theory is formulated in terms of classical commuting fields and provides symmetry-constrained diagnostics of marginality and finite-width effects through the maximal Lyapunov exponent, the amplification factor, and dressed spectral kernels. On top of this dynamics, we introduce a Markovian evolutionary scheme compatible with the local $U(1)$ structure of the effective model. By using a minimal implementation, the genotype is reduced to the weight-variance parameter $\sigma_w^2$, and the fitness functional combines spectral agreement, marginal stability, and a symmetry-constrained critical anchor. Comparing three evolutionary models, we find that only the fully symmetry-constrained Ginibre $U(1)$ version robustly approaches a narrow near-marginal regime and reproduces the predicted low-frequency finite-width spectral behavior. These results support the use of symmetry-guided effective stability diagnostics as practical principles for stochastic architecture search in controlled settings.
\end{abstract}

\maketitle

\section{Introduction}
\label{sec:intro}

Understanding how symmetry constrains stability and information propagation in deep neural networks has become an important theme in theoretically informed machine learning. In particular, the regime known as the edge of chaos has long been recognized as a distinguished operating point, where perturbations neither vanish too rapidly nor diverge across depth~\cite{Sompolinsky1988,poole2016exponential,Langton1990}. More broadly, recent work has shown that wide random networks admit mean-field and large-\(N\) descriptions closely connected to methods from statistical physics and quantum field theory~\cite{Roberts2022,Bondesan2021,Halverson2021,Erbin2022}. These developments suggest that symmetry principles and effective field-theoretic tools can play a useful role in organizing the dynamics of deep architectures.
In our previous published work~\cite{Terin2026}, we developed a gauge-covariant stochastic effective field theory for neural stability. That framework was formulated entirely in terms of classical commuting fields and combined three ingredients: a complex matter field representing coarse-grained feature amplitudes, a real Abelian connection field representing effective connectivity structure, and a fictitious stochastic depth variable governing noisy propagation. Its relation to Abelian gauge theory was structural rather than literal: local $U(1)$ covariance, covariant derivatives, gauge fixing, Ward-type constraints, and perturbative dressed-kernel reasoning provided the organizing language for the analysis. Within that setting, the edge of chaos was formulated as a linear-response marginality condition, and finite-width effects appeared as controlled deformations of the effective propagation kernels.

A complementary line of research has developed symmetry-equivariant and gauge-aware neural architectures directly at the architectural level~\cite{Cohen2016,Thomas2018,Fuchs2020,Finzi2021,Hutchinson2021,Theodosis2024}. These approaches encode symmetry by construction through equivariant layers, local transport rules, or loop-based features reminiscent of lattice gauge theories~\cite{PhysRevD.10.2445}. Their common message is that explicit symmetry constraints can improve robustness, inductive bias, and interpretability. The present work is not an alternative to that literature, but a continuation of it from the viewpoint of effective stochastic dynamics: rather than designing a single microscopic architecture, we ask how a symmetry-constrained effective theory can guide the search over architectures themselves.
This question naturally connects to neuroevolution. Evolutionary search methods, from classical schemes such as NEAT~\cite{Stanley2002} and CMA-ES~\cite{Hansen2003} to more recent quantum-inspired variants~\cite{Liu2019,Chen2020,Li2022}, provide flexible tools for exploring architecture spaces without relying exclusively on backpropagation. They are particularly attractive when the search space is discrete, highly nonconvex, or structured by hard constraints. At the same time, most such methods remain largely agnostic to the dynamical stability properties of the architectures they generate. In practice, architectures are explored by mutation and selection, but the search is rarely tied directly to symmetry-constrained criteria for marginality or finite-width stability.

Machine learning itself has also become an important tool for identifying phase transitions and critical behavior in classical and quantum systems~\cite{Ng2023UnsupervisedPT,Tian2023MLStructureProperty,VanNieuwenburg2017Confusion,Carrasquilla2017MLPhases,Ni2019DynamicalPT,Che2020TopologicalPT,Kuliashov2023DQPT,Tanaka2017CNNPT,Wang2016UnsupervisedPT}. In our own previous work on the classical Ising model~\cite{Terin2026IsingNAS}, neuroevolutionary neural architecture search was used to quantify how architectural and hyperparameter choices affect regression accuracy and the localization of critical points. That study showed that physical learning tasks can be highly sensitive to evolutionary design choices, but it did not address the dynamical organization of the search itself. In particular, it did not ask whether the search could be constrained by symmetry in such a way that near-critical behavior becomes a preferred attractor of the evolutionary dynamics.
The present paper addresses that question using a minimal setting. Building on our gauge-covariant effective theory of~\cite{Terin2026}, we promote architecture-level parameters to slow stochastic variables and formulate a symmetry-constrained evolutionary dynamics in function space. The resulting process is Markovian and admits a generator form analogous to drift-diffusion evolution on a structured manifold of effective models. The role of the local $U(1)$ structure is not to turn the architecture search into a literal quantum system, but to constrain the admissible directions of evolution and to preserve the same covariance principles that organize the underlying propagation dynamics.

In our implementation, we restrict attention to a minimal scalar genotype, the weight variance \(\sigma_w^2\). This choice is intentionally modest. It allows us to test whether the stability logic inherited from the effective field theory can act as a useful guide for stochastic architecture evolution before introducing more complicated architecture-level degrees of freedom. The key observables are the same as in the underlying effective theory: the maximal Lyapunov exponent, which diagnoses marginality, and the low-frequency power spectrum, which probes finite-width corrections to the dressed kernel.
This viewpoint also connects naturally with current discussions in mechanistic interpretability. Recent work has shown that strong structural constraints, such as explicit sparsity or circuit-level restrictions, can produce networks whose internal organization is substantially easier to analyze~\cite{Gao2025WeightSparseCircuits}. Our approach is complementary. Instead of imposing sparsity as the primary organizing principle, we impose a local covariance structure and use it to constrain both propagation and architecture evolution. In this sense, the framework developed here may be viewed as a symmetry-guided model organism: a restricted but analytically interpretable family in which stability and near-criticality arise from functional constraints rather than purely empirical tuning.

The contributions of this paper are therefore the following:
\begin{enumerate}[label=(\roman*)]
\item We extend our previos gauge-covariant stochastic neural-field framework of~\cite{Terin2026} by promoting architecture-level parameters to slow stochastic variables evolving in function space.
\item We formulate a symmetry-constrained Markovian evolutionary scheme whose drift-diffusion dynamics is compatible with the local $U(1)$ structure of the underlying effective model.
\item We define a practical minimal implementation in which the genotype is the scalar variance parameter \(\sigma_w^2\), and the fitness functional combines spectral agreement, marginal stability, and a symmetry-constrained critical anchor.
\item We compare three evolutionary implementations and show that only the fully symmetry-constrained Ginibre/$U(1)$ version robustly approaches a narrow near-marginal regime while reproducing the predicted low-frequency spectral behavior.
\item We interpret these results as evidence that the stability diagnostics derived from the effective field theory can serve as principled guides for stochastic architecture search in a controlled setting.
\end{enumerate}

This paper is organized as follows. Section~\ref{sec:ginn-recap} briefly reviews the gauge-covariant stochastic neural-field framework developed previously, including its effective action, MSRJD representation, stability criterion, and finite-width interpretation. Section~\ref{sec:evolution-space} introduces stochastic evolution in function space and formulates the symmetry-constrained search problem at the architecture level. Section~\ref{sec:evol-operators} defines the corresponding evolutionary generator and the fitness-driven dynamics used in practice. Section~\ref{sec:mqne-framework} presents the numerical implementation and compares three versions of the evolutionary scheme. Finally, Section~\ref{sec:conclusion} summarizes the main results and discusses directions for future work, including multi-parameter genotypes and extensions to richer architecture classes.

\section{Recap: Gauge-covariant stochastic neural fields}
\label{sec:ginn-recap}

This section summarizes the effective stochastic framework developed in our previous published work \cite{Terin2026}, in a form adapted to this manuscript. The purpose is to recall the minimal ingredients needed for the evolutionary layer introduced later: the gauge-covariant effective action, its stochastic MSRJD representation, the two-replica stability criterion, and the perturbative interpretation of finite-width effects.

Our formulation is built from classical commuting fields. The effective matter sector is described by a complex field $\phi(x,t)$ and its complex conjugate $\phi^\ast(x,t)$, while the effective connectivity sector is described by a real Abelian field $W_\mu(x,t)$. Here $x$ is an effective coordinate organizing feature, spatial, or latent directions depending on the architecture under consideration, and $t$ is a fictitious stochastic depth variable.
The local $U(1)$ structure is defined by
\begin{equation}
\phi(x,t)\rightarrow e^{i\theta(x,t)}\phi(x,t),\qquad
\phi^\ast(x,t)\rightarrow e^{-i\theta(x,t)}\phi^\ast(x,t),\qquad
W_\mu(x,t)\rightarrow W_\mu(x,t)-\frac{1}{g}\partial_\mu\theta(x,t),
\label{eq:u1-symmetry}
\end{equation}
so that the covariant derivative
\begin{equation}
D_\mu=\partial_\mu+i g W_\mu
\end{equation}
transforms in the standard Abelian way. The corresponding effective action is
\begin{equation}
S_{\mathrm{eff}}[\phi^\ast,\phi,W]
=
\int d^dx\,\Big[
(D_\mu\phi)^\ast(D_\mu\phi)
+m^2\phi^\ast\phi
+U(\phi^\ast\phi)
+\frac{1}{4}F_{\mu\nu}F_{\mu\nu}
+\frac{1}{2\alpha}(\partial_\mu W_\mu)^2
\Big],
\label{eq:ginnaction}
\end{equation}
with
\begin{equation}
F_{\mu\nu}=\partial_\mu W_\nu-\partial_\nu W_\mu.
\end{equation}
Equation~\eqref{eq:ginnaction} should be understood as an effective continuum model for neural propagation, not as a literal realization of quantum electrodynamics. Its relation to Abelian gauge theory is structural: both share local phase covariance, covariant derivatives, gauge fixing, and Ward-type constraints, but the present framework remains entirely classical and stochastic.
To model finite-width fluctuations and noisy propagation, the fields evolve according to It\^o--Langevin equations,
\begin{align}
\partial_t\phi(x,t)
&=
-\,\frac{\delta S_{\mathrm{eff}}}{\delta \phi^\ast(x,t)}
+\eta(x,t), \nonumber\\
\partial_t\phi^\ast(x,t)
&=
-\,\frac{\delta S_{\mathrm{eff}}}{\delta \phi(x,t)}
+\eta^\ast(x,t), \nonumber\\
\partial_t W_\mu(x,t)
&=
-\,\frac{\delta S_{\mathrm{eff}}}{\delta W_\mu(x,t)}
+\xi_\mu(x,t),
\label{eq:langevinrecap}
\end{align}
with Gaussian noises
\begin{align}
\langle \eta(x,t)\eta^\ast(x',t')\rangle
&=
2\kappa_\phi\,\delta^{(d)}(x-x')\delta(t-t'), \nonumber\\
\langle \xi_\mu(x,t)\xi_\nu(x',t')\rangle
&=
2\kappa_W\,\delta_{\mu\nu}\delta^{(d)}(x-x')\delta(t-t').
\end{align}
The stochastic dynamics can be written in Martin--Siggia--Rose--Janssen--de~Dominicis form by introducing response fields \(\tilde{\phi}\), \(\tilde{\phi}^\ast\), and \(\tilde{W}_\mu\). The resulting functional is
\begin{align}
S_{\mathrm{MSRJD}}
=
\int dt\,d^dx\,\Big[
&\tilde{\phi}^\ast
\Big(
\partial_t\phi+\frac{\delta S_{\mathrm{eff}}}{\delta\phi^\ast}
\Big)
+
\tilde{\phi}
\Big(
\partial_t\phi^\ast+\frac{\delta S_{\mathrm{eff}}}{\delta\phi}
\Big)
\nonumber\\
&+
\tilde{W}_\mu
\Big(
\partial_tW_\mu+\frac{\delta S_{\mathrm{eff}}}{\delta W_\mu}
\Big)
-\kappa_\phi\,\tilde{\phi}^\ast\tilde{\phi}
-\kappa_W\,\tilde{W}_\mu\tilde{W}_\mu
\Big].
\label{eq:MSRJD}
\end{align}
The generating functional
\begin{equation}
\mathcal{Z}
=
\int \mathcal{D}\Phi\,\mathcal{D}\tilde{\Phi}\;
e^{-S_{\mathrm{MSRJD}}[\Phi,\tilde{\Phi}]}
\end{equation}
defines the stochastic field theory underlying the stability analysis.
A basic consequence of the local $U(1)$ structure is a Ward-type identity relating matter and connectivity correlations. In the gauge-fixed theory, this identity constrains the longitudinal sector of the dressed kernels and ensures that gauge-equivalent parametrizations at fixed kernel geometry yield the same gauge-invariant observables. On the neural side, however, the parameter $\alpha$ also labels an effective kernel geometry, so statements of invariance must always be interpreted at fixed $\alpha$.

Stability is analyzed through a two-replica construction. Two copies of the same stochastic system evolve under the same noise realization but with slightly different initial conditions. If $\delta h_\ell$ denotes the replica separation at depth $\ell$, the empirical Lyapunov indicator is
\begin{equation}
\lambda_{\max}
=
\frac{1}{L}\sum_{\ell=0}^{L-1}
\log
\frac{\|\delta h_{\ell+1}\|}{\|\delta h_\ell\|}.
\label{eq:lambda-recap}
\end{equation}
The edge of chaos is identified with the marginal condition \(\lambda_{\max}=0\).

At mean-field level, the corresponding amplification factor takes the form
\begin{equation}
\chi_{\mathrm{MF}}(\sigma_w^2)
=
\sigma_w^2\,
\mathbb{E}_{Z\sim\mathcal{N}(0,1)}
\Big[
\varphi'(\sqrt{q_\star}Z)^2
\Big],
\label{eq:chi-recap}
\end{equation}
where \(q_\star\) is the variance fixed point. The mean-field critical condition is
\begin{equation}
\chi_{\mathrm{MF}}(\sigma_{w,c}^2)=1.
\end{equation}
In the effective field-theory language used here, this condition is the leading approximation to the full marginality criterion \(\chi=1\), where \(\chi\) denotes the dressed amplification factor in the dominant fluctuation channel.
For the linear stochastic effective sector considered in the spectral analysis, the frequency-space correlator is
\begin{equation}
X^{(0)}(\omega)
=
c(\omega)\,
\frac{\omega^2+\gamma^2}{\omega^2+\gamma^2-\sigma_w^2},
\qquad
c(\omega)=\frac{2\kappa}{\omega^2+\gamma^2}.
\label{eq:X0recap}
\end{equation}
Finite-width corrections appear as subleading deformations of the dressed kernel. In the linear sector, the leading correction scales as \(T/N\), giving
\begin{equation}
X(\omega)
=
X^{(0)}(\omega)
+
\frac{\gamma T}{N}\,X^{(1)}(\omega)
+
O\!\big((T/N)^2\big),
\label{eq:Xrecap}
\end{equation}
with
\begin{equation}
X^{(1)}(\omega)
=
\frac{\sigma_w^2}{2}\,
c(\omega)\,
\frac{\omega^2+\gamma^2}{(\omega^2+\gamma^2-\sigma_w^2)^2}.
\end{equation}
At the perturbative order considered, this correction deforms the spectral shape but does not shift the marginality condition, because its projection onto the critical mode vanishes. This is the restricted sense in which the critical point is symmetry protected in the effective theory.
The resulting dictionary is summarized schematically as follows:
\begin{center}
\renewcommand{\arraystretch}{1.1}
\begin{tabular}{l|l}
\textbf{Abelian gauge-theory language} & \textbf{Neural effective interpretation}\\
\hline
Gauge field \(A_\mu\) & Connectivity field \(W_\mu\)\\
Complex matter field \(\varphi\) & Feature field \(\phi\)\\
Covariant derivative \(D_\mu\) & Symmetry-constrained propagation\\
Gauge parameter \(\alpha\) & Effective kernel geometry\\
Noise amplitudes \(\kappa\) & Stochastic regularization scales\\
Ward-type identity & Constraint on longitudinal dressed kernels\\
Perturbative kernel dressing & Finite-width correction\\
Critical coupling & Critical gain / marginality threshold\\
\end{tabular}
\end{center}
This effective approach is the starting point for this work. In the next section, we extend it by promoting architecture-level parameters to slow stochastic variables and by introducing an evolutionary dynamics constrained by the same local symmetry structure.

\section{Evolution in function space: stochastic symmetry-constrained neuroevolution}
\label{sec:evolution-space}

The stochastic gauge-covariant framework reviewed in Sec.~\ref{sec:ginn-recap} describes propagation and stability within a fixed effective model. We now extend this setting by allowing architecture-level parameters to evolve on a slower timescale. The goal is to describe a symmetry-constrained search process in function space, built on top of the effective neural dynamics rather than replacing it.
Let
\begin{equation}
\Theta=\{W,b,\alpha,\Lambda,\dots\}
\end{equation}
denote a set of architecture-level parameters specifying an effective model within the gauge-covariant class. Each configuration \(\Theta\) defines a function \(f_\Theta:\mathbb{R}^n\to\mathbb{R}^m\). The corresponding function space may be equipped with a metric
\begin{equation}
g_{ij}(\Theta)=
\mathbb{E}_{x\sim p(x)}
\left[
\frac{\partial f_\Theta(x)}{\partial \Theta_i}
\frac{\partial f_\Theta(x)}{\partial \Theta_j}
\right],
\label{eq:metric}
\end{equation}
which provides a natural geometry on the manifold of architectures. Depending on context, this metric may be viewed as a Fisher-type metric or, more generally, as a local quadratic measure of sensitivity in function space.
Because the underlying effective model is locally gauge covariant, the induced geometry is naturally defined on equivalence classes of configurations related by local $U(1)$ reparametrizations. In this sense, the relevant search space is not the raw parameter space itself, but a symmetry-constrained manifold of effective models. The role of the evolutionary layer is therefore to explore this space while preserving the local covariance structure of the underlying dynamics.
At the coarse-grained level, we model architecture evolution as a stochastic process for \(\Theta\),
\begin{equation}
\dot{\Theta}_i
=
-\,g^{ij}\frac{\partial \mathcal{F}}{\partial \Theta_j}
+\zeta_i(t),
\label{eq:gradflow}
\end{equation}
where \(\mathcal{F}(\Theta)\) is an effective fitness functional and \(\zeta_i(t)\) is a stochastic term with covariance
\begin{equation}
\langle \zeta_i(t)\zeta_j(t')\rangle
=
2D\,g^{ij}\delta(t-t').
\end{equation}
Equation~\eqref{eq:gradflow} is the analogue, in architecture space, of the Langevin dynamics used earlier for field variables. It combines deterministic drift toward lower fitness with stochastic exploration along symmetry-compatible directions.
The corresponding probability density $P(\Theta,t)$ satisfies a Fokker--Planck equation,
\begin{equation}
\partial_t P
=
\nabla_i\!\left[
g^{ij}
\left(
\frac{\partial \mathcal{F}}{\partial \Theta_j}P
+
D\,\nabla_j P
\right)
\right],
\label{eq:FP}
\end{equation}
where $\nabla_i$ is the covariant derivative associated with the metric~\eqref{eq:metric}. This equation provides the basic Markovian description of stochastic architecture evolution.
For later purposes it is useful to rewrite Eq.~\eqref{eq:FP} in operator form. Introducing a density functional $\rho(\Theta,t)$, one may express the evolution schematically as
\begin{equation}
\partial_t \rho
=
\mathcal{L}_{\rm evo}\,\rho,
\label{eq:lindblad}
\end{equation}
where $\mathcal{L}_{\rm evo}$ is a linear generator acting on distributions over architecture space. In the present work, this operator language is used as a compact representation of the Markovian evolutionary dynamics. The analogy with Lindblad-type evolution is therefore structural: it emphasizes the coexistence of drift and diffusion in a generator form, but no genuinely quantum degree of freedom is introduced in the architecture sector.
Discretizing the evolution in time, one obtains a Markov chain over effective models,
\begin{equation}
\rho_{k+1}
=
\mathcal{E}_{k,k+1}[\rho_k],
\label{eq:markov}
\end{equation}
where \(\mathcal{E}_{k,k+1}\) denotes the transition map from one generation to the next. In practice, \(\mathcal{E}_{k,k+1}\) combines three ingredients:
\begin{enumerate}[label=(\roman*)]
\item a deterministic tendency toward lower fitness,
\item a stochastic mutation step,
\item a projection onto the symmetry-compatible model class.
\end{enumerate}
The local $U(1)$ structure imposes a covariance condition on this map:
\begin{equation}
\mathcal{E}_{k,k+1}[G\rho G^{-1}]
=
G\,\mathcal{E}_{k,k+1}[\rho]\,G^{-1},
\label{eq:gauge-condition}
\end{equation}
where $G$ denotes the action of the local symmetry on the effective model variables. Equation~\eqref{eq:gauge-condition} states that evolutionary updates must respect the same covariance structure that governs the underlying field dynamics. This is the defining symmetry constraint of the present neuroevolutionary scheme.
It is useful to write the architecture evolution in a path-integral form analogous to the MSRJD construction of the field sector. Introducing response variables $\hat{\Theta}_i$, the generating functional becomes
\begin{equation}
\mathcal{Z}_{\mathrm{evo}}
=
\int
\mathcal{D}\Theta\,\mathcal{D}\hat{\Theta}\;
\exp\!\Big[
-\!\int\!dt\,
\Big(
\hat{\Theta}_i\,\dot{\Theta}_i
+
\hat{\Theta}_i g^{ij}\frac{\partial\mathcal{F}}{\partial\Theta_j}
-
D\,\hat{\Theta}_i g^{ij}\hat{\Theta}_j
\Big)
\Big].
\label{eq:EMSJRD}
\end{equation}
This functional should be read as the stochastic representation of the evolutionary process in architecture space. It is the natural counterpart, at the macroscopic level, of the MSRJD functional governing the microscopic propagation dynamics.
The result is a two-layer stochastic description:
\begin{enumerate}[label=(\roman*)]
\item the inner layer describes field propagation and stability for fixed effective parameters;
\item the outer layer describes slow stochastic evolution of those parameters across the symmetry-constrained architecture manifold.
\end{enumerate}
The same local covariance structure underlies both levels.
This hierarchy is the basic motivation for the present framework. Rather than treating architecture search as an external optimization procedure unrelated to the dynamical theory, we embed it into a stochastic evolution compatible with the symmetry structure of the effective model itself. In this way, architecture evolution inherits the same constraints that organize propagation, stability, and perturbative corrections in the underlying gauge-covariant field description.
In the next section we make this construction more explicit by introducing the corresponding symmetry-compatible evolutionary operators and the associated fitness functional used in the numerical implementation.

\section{Symmetry-compatible evolutionary operators and fitness-driven dynamics}
\label{sec:evol-operators}

The stochastic evolution introduced in Sec.~\ref{sec:evolution-space} can be expressed in terms of transition operators acting on distributions over the symmetry-constrained architecture manifold. In this section we make that structure explicit and define the fitness functional used in the numerical implementation.
Let $\rho(\Theta,t)$ denote the distribution of effective architectures at generation time $t$. Its evolution is governed by a linear generator
\begin{equation}
\partial_t \rho
=
\mathcal{L}_{\mathrm{evo}}\,\rho,
\label{eq:lindblad-gen}
\end{equation}
where $\mathcal{L}_{\mathrm{evo}}$ contains both deterministic drift and stochastic exploration. At the level of Eq.~\eqref{eq:FP}, this generator is simply the Fokker--Planck operator associated with the fitness landscape and the metric on function space.
A convenient decomposition is
\begin{equation}
\mathcal{L}_{\mathrm{evo}}
=
\mathcal{L}_{\mathrm{drift}}
+
\mathcal{L}_{\mathrm{diff}},
\end{equation}
with
\begin{equation}
\mathcal{L}_{\mathrm{drift}}\rho
=
\nabla_i\!\left(
g^{ij}\frac{\partial\mathcal{F}}{\partial\Theta_j}\rho
\right),
\qquad
\mathcal{L}_{\mathrm{diff}}\rho
=
D\,\nabla_i\!\left(g^{ij}\nabla_j\rho\right).
\end{equation}
The first term drives the population toward lower fitness, while the second term maintains exploration over the effective architecture space.
The local $U(1)$ structure of the underlying neural field model imposes a covariance requirement on the evolutionary dynamics. If $G(\theta)$ denotes the local symmetry action on the effective variables, the generator must satisfy
\begin{equation}
\mathcal{L}_{\mathrm{evo}}\!\left[G\rho G^{-1}\right]
=
G\,\mathcal{L}_{\mathrm{evo}}[\rho]\,G^{-1}.
\label{eq:gauge-commutation}
\end{equation}
This condition ensures that evolution preserves the covariance structure of the underlying model and does not generate motion along unphysical reparametrization directions.
In discrete time, the same statement is expressed by the transition map
\begin{equation}
\rho_{k+1}
=
\mathcal{E}_{k,k+1}[\rho_k],
\label{eq:rho-evol}
\end{equation}
with covariance condition
\begin{equation}
\mathcal{E}_{k,k+1}[G\rho G^{-1}]
=
G\,\mathcal{E}_{k,k+1}[\rho]\,G^{-1}.
\label{eq:propagator-gauge}
\end{equation}
Operationally, $\mathcal{E}_{k,k+1}$ combines selection, mutation, and projection onto the admissible model class.
For the practical implementation used here, we restrict the genotype to a single scalar control parameter, the weight variance $\sigma_w^2$. The corresponding fitness functional penalizes three effects:
\begin{enumerate}[label=(\roman*)]
\item mismatch between the simulated and theoretical low-frequency spectra,
\item departure from marginality as measured by the Lyapunov exponent,
\item departure from the symmetry-constrained critical reference value.
\end{enumerate}
Accordingly, we define
\begin{equation}
\mathcal{F} =
w_{\mathrm{spec}}\,
\mathrm{RelMSE}_{\omega<2}\!\big[X_{\mathrm{sim}},X_{\mathrm{th}}\big]
+ w_{\lambda}\,\lambda_{\max}^2
+ w_{\mathrm{crit}}\,\big(\sigma_w^2-\gamma^2\big)^2.
\label{eq:Vterms}
\end{equation}
Here $\mathrm{RelMSE}_{\omega<2}$ denotes the relative mean-squared error between the simulated and theoretical spectra in the low-frequency band, $\lambda_{\max}$ is the empirical instability indicator, and the last term anchors the search near the symmetry-constrained critical reference value $\sigma_w^2=\gamma^2$. This last ingredient should not be interpreted as a universal identity valid across arbitrary kernel families, but as the critical reference inherited from the effective model class considered in this work.
The selection step is implemented through Boltzmann weights,
\begin{equation}
p(\Theta)\propto e^{-\beta \mathcal{F}(\Theta)},
\end{equation}
with an annealed inverse temperature \(\beta\). Mutation is modeled by Gaussian perturbations of the genotype, followed by projection back into the admissible interval. In this way, the numerical scheme realizes a concrete stochastic search process constrained by the same symmetry logic that governs the effective field theory.
Although the current implementation evolves only the scalar genotype $\sigma_w^2$, the operator formalism above makes clear how the construction extends to higher-dimensional architecture spaces. More general choices of $\Theta$ would simply enlarge the state space on which $\mathcal{L}_{\mathrm{evo}}$ acts, while preserving the same drift-diffusion structure and covariance constraints.

The practical role of the present section is therefore twofold. First, it translates the abstract evolution-in-function-space idea into a concrete Markovian search process. Second, it shows how the stability diagnostics derived from the effective field theory enter directly into the fitness landscape explored by the algorithm.
In the next section we show the numerical implementation and compare three versions of the stochastic evolutionary scheme, highlighting the role of the symmetry-constrained critical anchor in the resulting dynamics.

\section{Simulation framework, models, and results}
\label{sec:mqne-framework}

We now display the numerical implementation of the symmetry-constrained stochastic evolutionary scheme. The goal is to test, in a minimal setting, whether the stability diagnostics inherited from the gauge-covariant effective theory can guide architecture evolution toward a marginal regime.
In our current implementation, the genotype is reduced to a single scalar parameter, the weight variance $\sigma_w^2$. This is a minimal choice that enables us to isolate the role of the stability criterion without introducing additional architectural degrees of freedom. The neural dynamics is taken to be the linear stochastic model
\begin{equation}
h_{t+1}
=
h_t
+
\Delta t\,
\big[-\gamma h_t + W h_t\big]
+
\sqrt{2\kappa\Delta t}\,\eta_t,
\end{equation}
where $W$ is drawn from a real Ginibre ensemble with variance $\sigma_w^2/N$, and $\eta_t$ is Gaussian noise. In the gauge-covariant version, we also include local $U(1)$-type phases through the parametrization
\begin{equation}
W_{ij}=|W_{ij}|\cos\theta_{ij},
\qquad
\theta_{ij}\sim\mathcal{N}(0,\mathrm{phase\_std}^2),
\end{equation}
which preserves the symmetry-constrained interpretation of the effective propagation sector.
For this linear drift, the largest Lyapunov exponent is estimated from the spectral abscissa of $W-\gamma I$, while the power spectral density $X_{\mathrm{sim}}(\omega)$ is compared with the effective prediction
\begin{equation}
X_{\mathrm{th}}(\omega)
=
X^{(0)}(\omega)+\frac{\gamma T}{N}X^{(1)}(\omega),
\end{equation}
derived in the previous sections.
At each generation, the spectrum is averaged over several seeds and the fitness functional
\begin{equation}
\mathcal{F} =
w_{\mathrm{spec}}\,
\mathrm{RelMSE}_{\omega<2}\!\big[X_{\mathrm{sim}},X_{\mathrm{th}}\big]
+ w_{\lambda}\,\lambda_{\max}^2
+ w_{\mathrm{crit}}\,\big(\sigma_w^2-\gamma^2\big)^2
\end{equation}
is evaluated. Selection is Boltzmann weighted,
\begin{equation}
p(\sigma_w^2)\propto e^{-\beta \mathcal{F}},
\end{equation}
with annealed \(\beta\), and mutations are implemented as Gaussian perturbations followed by clipping to the admissible interval.
The reference hyperparameters used in the simulations are:
$N=256$, $L=2000$, $\Delta t=0.05$, $\gamma=\kappa=1$, population size $P=48$, number of generations $K=100$, initial distribution $\sigma_w^2\sim\mathcal{U}(0.30,1.30)$, mutation schedule $\mathrm{mut_std}=0.02\times 0.98^k$, and local phases with $\mathrm{phase_std}=0.3$. The ratio $T/N$ is fixed by $T=L\Delta t$, giving $T/N\simeq 100/256\approx 0.391$.
We then compare three implementations.

\paragraph{Model A: baseline without critical anchor.}

In Model A, the critical anchoring term is removed from the fitness, so that selection is driven only by spectral agreement and the Lyapunov penalty. In this case, the population drifts toward subcritical variances and remains in the ordered regime, with $\lambda_{\max}<0$ throughout evolution. The resulting spectra reproduce the general Ornstein--Uhlenbeck background at high frequency but suppress low-frequency power. This model shows that stochastic search alone does not maintain marginality.

\paragraph{Model B: real-symmetric critical anchor.}

In Model B, the fitness includes a critical anchor, but the connectivity ensemble is restricted to a real-symmetric setting analogous to a GOE-type reference. The population stabilizes closer to marginality and $\lambda_{\max}$ fluctuates around zero, but the resulting dynamics remains more rigid than in the full gauge-covariant case. Spectral agreement improves in the low-frequency regime, yet the restriction to purely real symmetric fluctuations limits the effective exploration of the propagation sector.

\paragraph{Model C: symmetry-constrained Ginibre/$U(1)$ implementation.}

Model C is the main implementation. Here we combine the critical anchor with the real Ginibre ensemble and local $U(1)$-type phases. In this case, the population self-organizes toward a narrow band around $\lambda_{\max}\approx 0$, without manual tuning of $\sigma_w^2$. The best individuals lie close to the critical reference value, and the averaged spectrum matches the predicted form $X_{\mathrm{th}}(\omega)$ in the low- and intermediate-frequency regimes, including the leading $T/N$ correction.
\begin{figure}
\centering
\begin{subfigure}{0.48\textwidth}
  \centering
  \includegraphics[width=\linewidth]{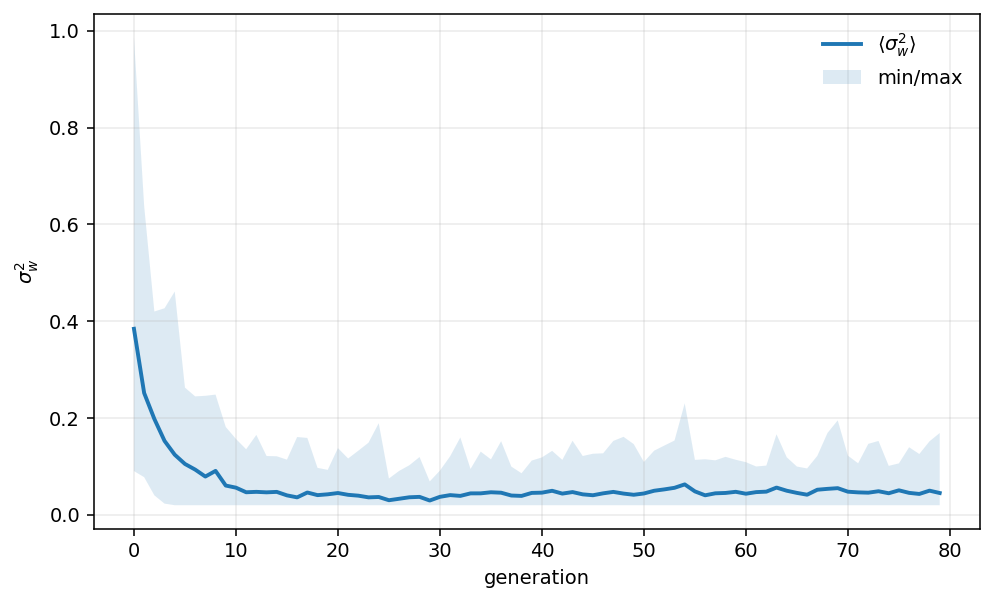}
  \caption{Model A: baseline without critical anchor.}
  \label{fig:sigma_A}
\end{subfigure}
\hfill
\begin{subfigure}[t]{0.48\textwidth}
  \centering
  \includegraphics[width=\linewidth]{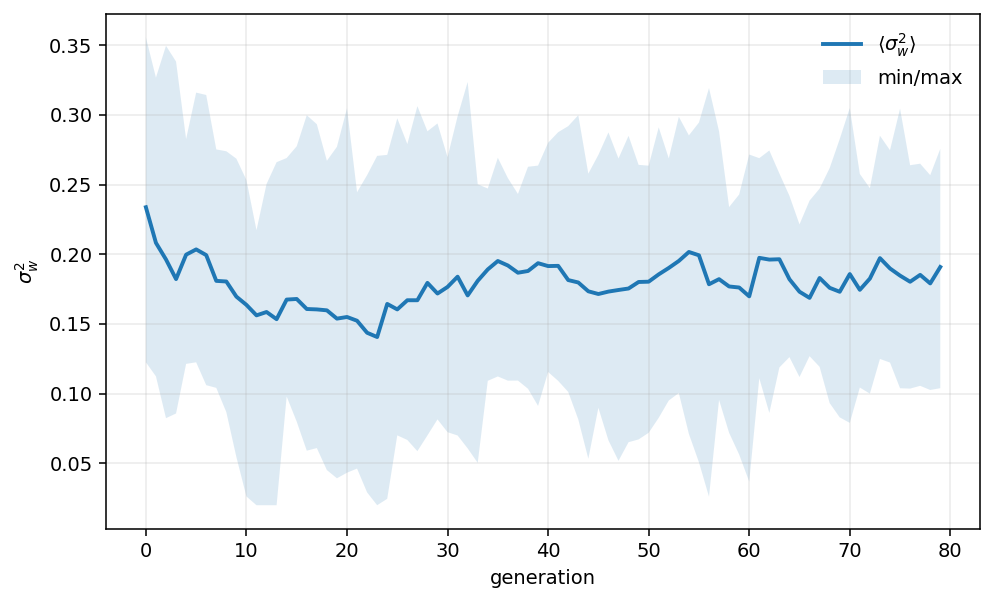}
  \caption{Model B: real-symmetric anchored evolution.}
  \label{fig:sigma_B}
\end{subfigure}

\vspace{1em}

\begin{subfigure}{0.68\textwidth}
  \centering
  \includegraphics[width=\linewidth]{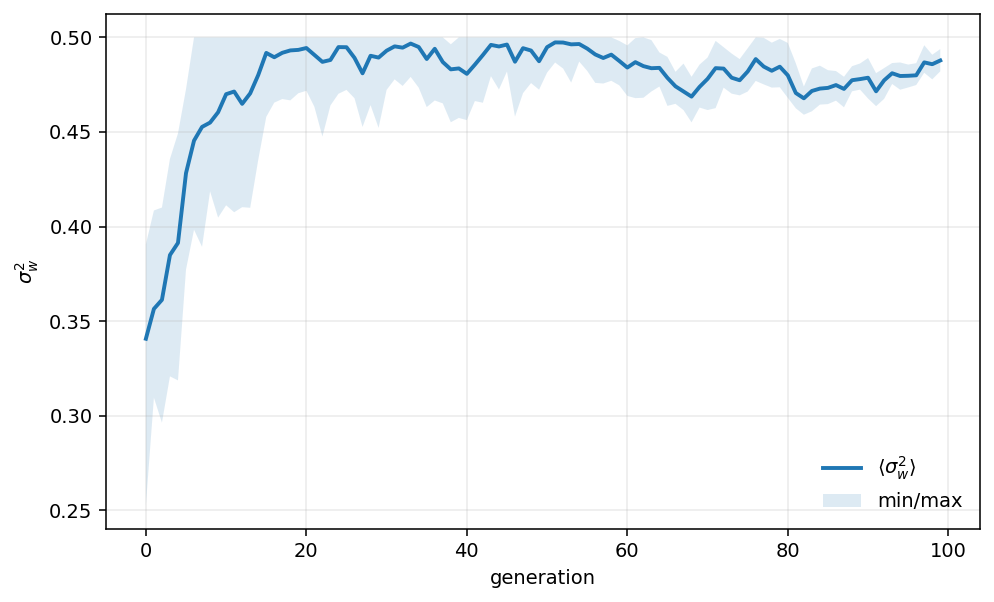}
  \caption{Model C: symmetry-constrained Ginibre/$U(1)$ evolution.}
  \label{fig:sigma_C}
\end{subfigure}

\caption{Evolution of the weight variance \(\sigma_w^2\) for the three stochastic evolutionary schemes. Model C remains closest to the symmetry-constrained critical reference value.}
\label{fig:sigma_models}
\end{figure}

\begin{figure}
\centering
\begin{subfigure}[t]{0.48\textwidth}
  \centering
  \includegraphics[width=\linewidth]{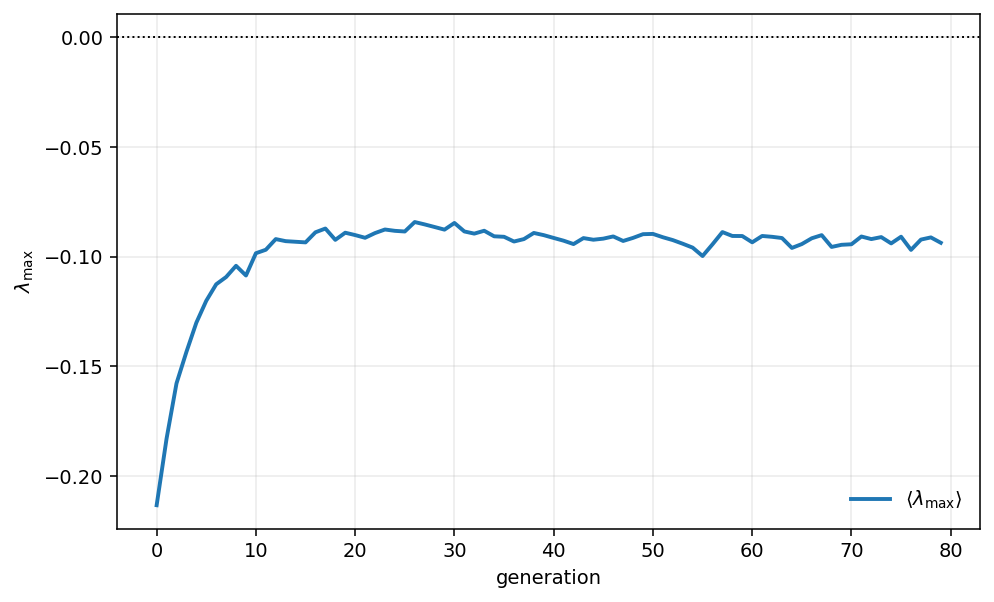}
  \caption{Model A: ordered regime.}
  \label{fig:lambda_A}
\end{subfigure}
\hfill
\begin{subfigure}{0.48\textwidth}
  \centering
  \includegraphics[width=\linewidth]{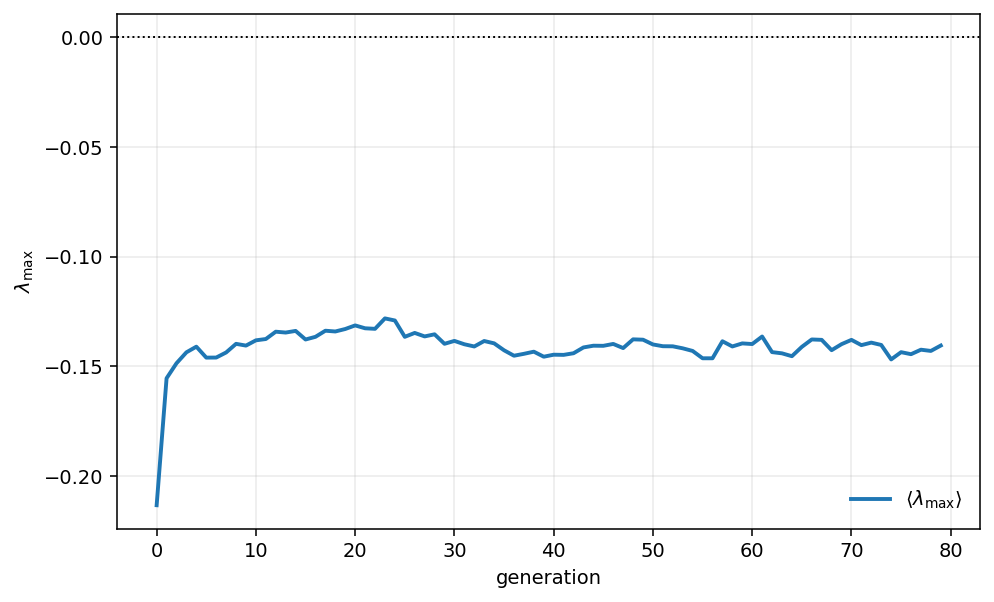}
  \caption{Model B: near-marginal fluctuations.}
  \label{fig:lambda_B}
\end{subfigure}

\vspace{1em}

\begin{subfigure}{0.68\textwidth}
  \centering
  \includegraphics[width=\linewidth]{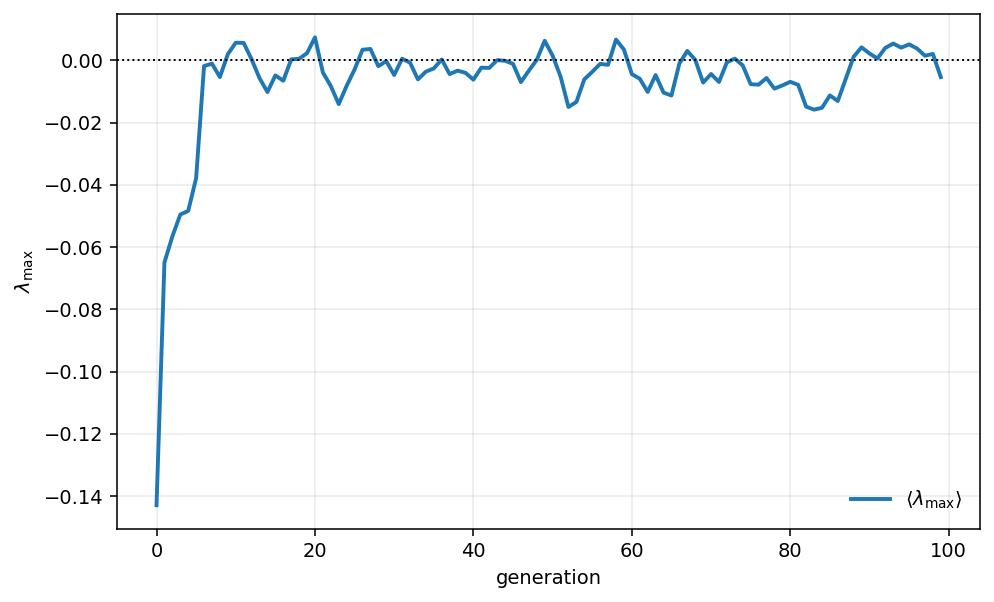}
  \caption{Model C: approach to the marginal regime.}
  \label{fig:lambda_C}
\end{subfigure}

\caption{Evolution of the largest Lyapunov exponent \(\lambda_{\max}\). Only Model C robustly approaches the marginal regime \(\lambda_{\max}\approx 0\).}
\label{fig:lambda_models}
\end{figure}

\begin{figure}
\centering
\begin{subfigure}{0.48\textwidth}
  \centering
  \includegraphics[width=\linewidth]{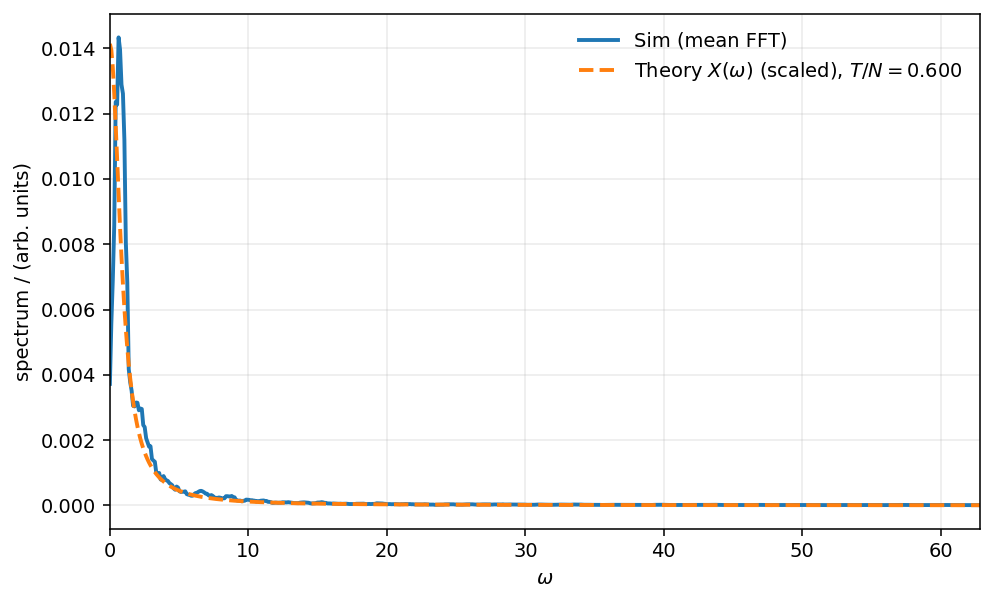}
  \caption{Model A.}
  \label{fig:psd_A}
\end{subfigure}
\hfill
\begin{subfigure}[t]{0.48\textwidth}
  \centering
  \includegraphics[width=\linewidth]{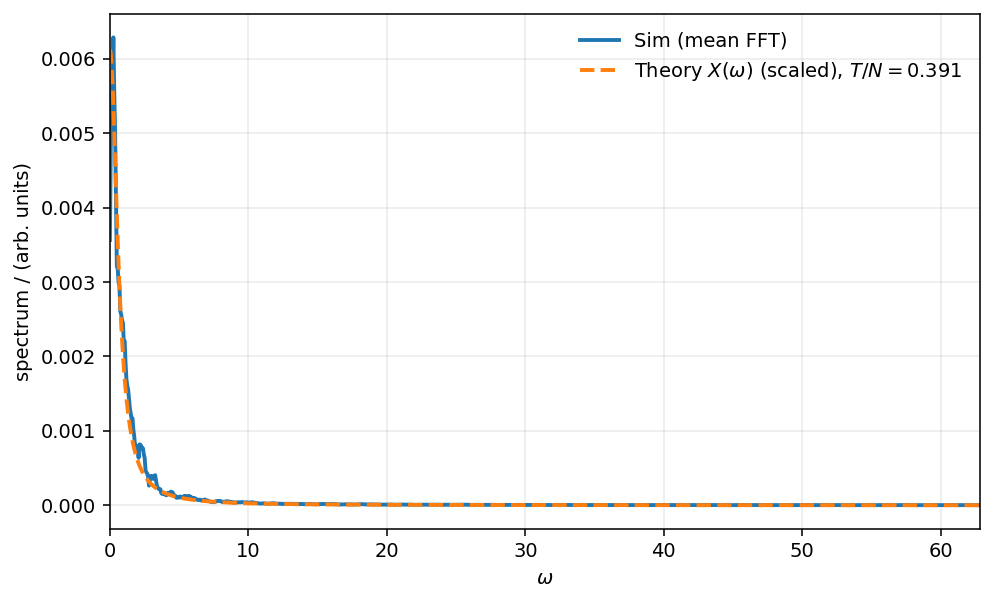}
  \caption{Model B.}
  \label{fig:psd_B}
\end{subfigure}

\vspace{1em}

\begin{subfigure}{0.68\textwidth}
  \centering
  \includegraphics[width=\linewidth]{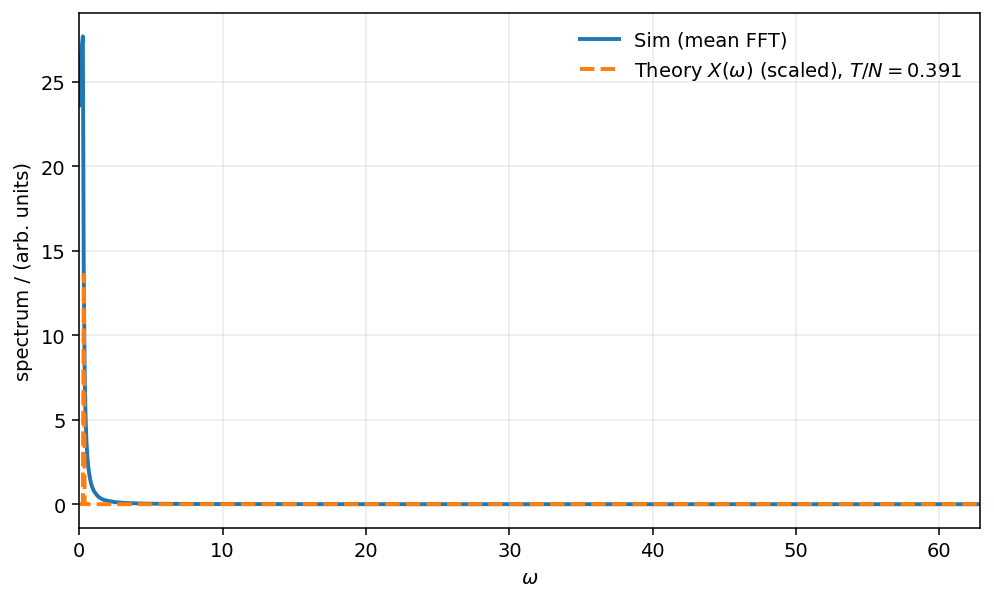}
  \caption{Model C.}
  \label{fig:psd_C}
\end{subfigure}

\caption{Power spectra for the three evolutionary schemes. Model C yields the closest agreement with the effective theoretical prediction, including the leading finite-width correction.}
\label{fig:psd_models}
\end{figure}
%\newpage
These three implementations illustrate the role of the symmetry-constrained critical anchor. Without it, the search drifts toward the ordered phase. With a partial anchoring in a more restricted ensemble, the dynamics approaches marginality but remains less flexible. In the full symmetry-constrained implementation, the search consistently approaches the marginal regime while preserving agreement with the effective spectral prediction.
The main conclusion of this section is therefore limited but clear: within the minimal scalar-genotype implementation studied here, symmetry-constrained stochastic evolution provides a practical mechanism for steering the system toward a marginally stable regime. The results do not establish a universal theorem for arbitrary architecture spaces, but they do show that the stability logic derived from the effective field theory can successfully guide a nontrivial evolutionary search.
\newpage

\section{Conclusion and outlook}
\label{sec:conclusion}

We have developed a symmetry-constrained stochastic neuroevolution framework built on top of the gauge-covariant effective neural-field model introduced in our previous published work \cite{Terin2026}. Our construction combines two levels of dynamics: a microscopic stochastic field description of propagation and stability, and a macroscopic stochastic evolution of architecture-level parameters in function space. In the present implementation, this second layer is realized through a Markovian evolutionary process acting on a minimal scalar genotype, the weight variance $\sigma_w^2$.
Moreover, the fundamental effective theory remains entirely classical and stochastic. Its relation to Abelian gauge theory is structural: local $U(1)$ covariance, covariant kernels, gauge fixing, Ward-type constraints, and perturbative dressed-kernel reasoning provide the organizing principles of the analysis. Within our framework, the edge of chaos is identified with a marginality condition, $\lambda_{\max}=0$, equivalently $\chi=1$, and finite-width effects appear as controlled deformations of the relevant propagation kernels.

Our numerical study shows that this stability logic can be incorporated into an evolutionary search scheme. In the baseline model without critical anchoring, the search drifts toward the ordered regime. In the anchored real-symmetric model, the dynamics approaches marginality but remains comparatively restricted. In the symmetry-constrained Ginibre/$U(1)$ implementation, the search robustly approaches a narrow near-marginal regime and reproduces the predicted low-frequency spectral behavior, including the leading finite-width correction. Taken together, these results show that the effective stability criterion can serve as a practical guide for stochastic architecture evolution.
From a physical viewpoint, our approach may be read as a two-layer stochastic field system in which the same local covariance structure constrains both propagation and search. From a computational viewpoint, it provides a symmetry-guided alternative to purely heuristic architecture tuning: instead of relying only on ad hoc regularization or manual critical initialization, the search is biased toward a marginally stable regime by construction.

At the same time, our current implementation explores only a one-parameter genotype and uses a linear stochastic effective sector for the spectral analysis. Our conclusions should therefore be interpreted as evidence that symmetry-constrained stochastic evolution is viable and informative in a controlled setting, not as a universal theorem about arbitrary neural architecture spaces or nonlinear models.
Several directions remain open. On the theoretical side, it would be natural to extend the architecture dynamics to genuinely multi-parameter genotypes, to analyze higher-order corrections to the critical surface, and to clarify how different kernel geometries define distinct universality classes of propagation. On the modeling side, it remains to be studied how the present approach adapts to richer classes of architectures, including convolutional, graph-based, and explicitly equivariant networks.
Our main conclusion is therefore precise: a gauge-covariant effective description of neural stability can be combined with a stochastic evolutionary layer to produce a symmetry-constrained architecture search framework, in which marginality and finite-width effects are not imposed heuristically but organized by the same functional structure that governs the fundamental dynamics.

\bibliographystyle{unsrtnat}
\bibliography{refs}

@article{Terin2026,
  author  = {Terin, Rodrigo Carmo},
  title   = {Gauge-covariant stochastic neural fields: stability and finite-width effects},
  journal = {Scientific Reports},
  year    = {2026},
  issn    = {2045-2322},
  doi     = {10.1038/s41598-026-47071-y}
}

@article{PhysRevD.10.2445,
  title = {Confinement of quarks},
  author = {Wilson, Kenneth G.},
  journal = {Phys. Rev. D},
  volume = {10},
  issue = {8},
  pages = {2445--2459},
  numpages = {0},
  year = {1974},
  month = {Oct},
  publisher = {American Physical Society},
  doi = {10.1103/PhysRevD.10.2445},
  url = {https://link.aps.org/doi/10.1103/PhysRevD.10.2445}
}

@misc{Gao2025WeightSparseCircuits,
  author       = {Leo Gao and Achyuta Rajaram and Jacob Coxon and
                  Soham V. Govande and Bowen Baker and Dan Mossing},
  title        = {Weight-sparse transformers have interpretable circuits},
  year         = {2025},
  howpublished = {\url{https://cdn.openai.com/pdf/41df8f28-d4ef-43e9-aed2-823f9393e470/circuit-sparsity-paper.pdf}},
  note         = {OpenAI technical report},
}

@article{Sompolinsky1988,
  author={H. Sompolinsky and A. Crisanti and H.-J. Sommers},
  title={Chaos in random neural networks},
  journal={Phys. Rev. Lett.},
  volume={61},
  pages={259},
  year={1988}
}

@inproceedings{poole2016exponential,
  author={B. Poole and S. Lahiri and M. Raghu and J. Sohl-Dickstein and S. Ganguli},
  title={Exponential expressivity in deep neural networks through transient chaos},
  booktitle={Advances in Neural Information Processing Systems 29},
  pages={3360--3368},
  year={2016}
}

@article{Langton1990,
  author={C. G. Langton},
  title={Computation at the edge of chaos: phase transitions and emergent computation},
  journal={Physica D},
  volume={42},
  pages={12--37},
  year={1990}
}

@incollection{Terin2026IsingNAS,
  author    = {Rodrigo Carmo Terin and Zochil Gonzalez Arenas and Roberto Santana},
  title     = {Identifying Phase Transitions in the Classical Ising Model with Neural Networks: A Neural Architecture Search Perspective},
  booktitle = {Pattern Recognition and Artificial Intelligence},
  series    = {Lecture Notes in Networks and Systems},
  volume    = {1393},
  publisher = {Springer},
  address   = {Cham},
  year      = {2026},
  doi       = {10.1007/978-3-031-90893-4_50},
  isbn      = {978-3-031-90892-7},
  note      = {Online ISBN: 978-3-031-90893-4}
}

@article{Ng2023UnsupervisedPT,
  author  = {K.-K. Ng and M.-F. Yang},
  title   = {Unsupervised learning of phase transitions via modified anomaly detection with autoencoders},
  journal = {Physical Review B},
  volume  = {108},
  number  = {21},
  pages   = {214428},
  year    = {2023}
}

@article{Tian2023MLStructureProperty,
  author  = {Z. Tian and S. Zhang and G.-W. Chern},
  title   = {Machine learning for structure-property relationships: Scalability and limitations},
  journal = {arXiv preprint},
  eprint  = {2304.05502},
  archivePrefix = {arXiv},
  primaryClass  = {cond-mat.mtrl-sci},
  year    = {2023}
}

@article{VanNieuwenburg2017Confusion,
  author  = {E. P. Van Nieuwenburg and Y.-H. Liu and S. D. Huber},
  title   = {Learning phase transitions by confusion},
  journal = {Nature Physics},
  volume  = {13},
  number  = {5},
  pages   = {435--439},
  year    = {2017}
}

@article{Carrasquilla2017MLPhases,
  author  = {J. Carrasquilla and R. G. Melko},
  title   = {Machine learning phases of matter},
  journal = {Nature Physics},
  volume  = {13},
  number  = {5},
  pages   = {431--434},
  year    = {2017}
}

@article{Ni2019DynamicalPT,
  author  = {Q. Ni and M. Tang and Y. Liu and Y.-C. Lai},
  title   = {Machine learning dynamical phase transitions in complex networks},
  journal = {Physical Review E},
  volume  = {100},
  pages   = {052312},
  year    = {2019}
}

@article{Che2020TopologicalPT,
  author  = {Y. Che and C. Gneiting and T. Liu and F. Nori},
  title   = {Topological quantum phase transitions retrieved through unsupervised machine learning},
  journal = {Physical Review B},
  volume  = {102},
  pages   = {134213},
  year    = {2020}
}

@article{Kuliashov2023DQPT,
  author  = {O. N. Kuliashov and A. A. Markov and A. N. Rubtsov},
  title   = {Dynamical quantum phase transition without an order parameter},
  journal = {Physical Review B},
  volume  = {107},
  pages   = {094304},
  year    = {2023}
}

@article{Tanaka2017CNNPT,
  author  = {A. Tanaka and A. Tomiya},
  title   = {Detection of phase transition via convolutional neural networks},
  journal = {Journal of the Physical Society of Japan},
  volume  = {86},
  number  = {6},
  pages   = {063001},
  year    = {2017}
}

@article{Wang2016UnsupervisedPT,
  author  = {L. Wang},
  title   = {Discovering phase transitions with unsupervised learning},
  journal = {Physical Review B},
  volume  = {94},
  number  = {19},
  pages   = {195105},
  year    = {2016}
}

@book{Roberts2022,
  author={D. A. Roberts and S. Yaida and B. Hanin},
  title={The Principles of Deep Learning Theory},
  publisher={Cambridge University Press},
  year={2022}
}

@inproceedings{Bondesan2021,
  author={R. Bondesan and M. Welling},
  title={The hintons in your neural network: a quantum field theory view of deep learning},
  booktitle={Proc. 38th Int. Conf. on Machine Learning},
  pages={1038--1048},
  year={2021}
}

@article{Halverson2021,
  author={J. Halverson and A. Maiti and K. Stoner},
  title={Neural networks and quantum field theory},
  journal={Mach. Learn.: Sci. Technol.},
  volume={2},
  pages={035002},
  year={2021}
}

@article{Erbin2022,
  author={H. Erbin and V. Lahoche and D. Samary},
  title={Nonperturbative renormalization for the neural network–QFT correspondence},
  journal={Mach. Learn.: Sci. Technol.},
  volume={3},
  pages={015027},
  year={2022}
}

@inproceedings{Cohen2016,
  author={T. S. Cohen and M. Welling},
  title={Group equivariant convolutional networks},
  booktitle={Proc. 33rd Int. Conf. on Machine Learning},
  pages={2990--2999},
  year={2016}
}

@inproceedings{Thomas2018,
  author={N. Thomas et al.},
  title={Tensor field networks: Rotation- and translation-equivariant neural networks for 3D point clouds},
  booktitle={arXiv:1802.08219},
  year={2018}
}

@article{Fuchs2020,
  author={F. Fuchs, D. Worrall, V. Fischer and M. Welling},
  title={SE(3)-Transformers: 3D Roto-Translation Equivariant Attention Networks},
  journal={Adv. Neural Inf. Process. Syst.},
  volume={33},
  pages={1970--1981},
  year={2020}
}

@inproceedings{Finzi2021,
  author={M. Finzi et al.},
  title={A Practical Method for Constructing Equivariant Multilayer Perceptrons for Arbitrary Matrix Groups},
  booktitle={ICML},
  year={2021}
}

@inproceedings{Hutchinson2021,
  author={M. Hutchinson, M. Finzi, E. Foerster and A. Wilson},
  title={LIENN: E(n)-Equivariant Graph Neural Networks},
  booktitle={NeurIPS},
  year={2021}
}

@inproceedings{Theodosis2024,
  author={M. Theodosis and D. E. Ba and N. Dehmamy},
  title={Constructing gauge-invariant neural networks for scientific applications},
  booktitle={ICML AI4Science Workshop},
  year={2024}
}

@inproceedings{Stanley2002,
  author={K. O. Stanley and R. Miikkulainen},
  title={Evolving neural networks through augmenting topologies},
  booktitle={Evolutionary Computation},
  year={2002}
}

@article{Hansen2003,
  author={N. Hansen and A. Ostermeier},
  title={Completely derandomized self-adaptation in evolution strategies},
  journal={Evol. Comput.},
  volume={9},
  pages={159--195},
  year={2003}
}

@article{Liu2019,
  author={J. Liu and J. Wang},
  title={Quantum-inspired evolutionary algorithm with density matrix encoding},
  journal={Inf. Sci.},
  volume={505},
  pages={178--197},
  year={2019}
}

@article{Chen2020,
  author={Y. Chen and Z. Li and S. Lin},
  title={Quantum-inspired multiobjective evolutionary algorithm},
  journal={IEEE Trans. Evol. Comput.},
  volume={24},
  pages={471--484},
  year={2020}
}

@article{Li2022,
  author={Y. Li and Q. Zhang and M. Gong},
  title={Quantum-inspired evolutionary computation: A survey},
  journal={ACM Comput. Surveys},
  volume={54},
  pages={1--38},
  year={2022}
}
\end{document}